\documentclass[conference,letterpaper]{IEEEtran}

\IEEEoverridecommandlockouts
\usepackage{cite}
\usepackage{amsmath,amssymb,amsfonts}
\usepackage{algorithmic}
\usepackage{graphicx}
\usepackage{textcomp}
\usepackage{xcolor}
\usepackage[switch]{lineno}
\usepackage{booktabs} % for tables
\usepackage{subfigure}

\def\BibTeX{{\rm B\kern-.05em{\sc i\kern-.025em b}\kern-.08em
    T\kern-.1667em\lower.7ex\hbox{E}\kern-.125emX}}
\begin{document}

 %\linenumbers
% %\switchlinenumbers
 
\title{Fully-automated patient-level malaria assessment on field-prepared thin blood film microscopy images, including Supplementary Information 
\thanks{*Equal contributions\newline
$~~~$We gratefully acknowledge support from the Bill and Melinda Gates Foundation Trust, through the Global Good Fund.}
}

\author{\IEEEauthorblockN{ Charles B. Delahunt\textsuperscript{1*}, Mayoore S. Jaiswal\textsuperscript{2*}, Matthew P. Horning\textsuperscript{1}, Samantha Janko\textsuperscript{3}, Clay M. Thompson\textsuperscript{4},\\
Sourabh Kulhare\textsuperscript{1}, Liming Hu\textsuperscript{1},  Travis Ostbye\textsuperscript{1}, Grace Yun\textsuperscript{1}, Roman Gebrehiwot\textsuperscript{1}, Benjamin K. Wilson\textsuperscript{1}, \\ 
Earl Long\textsuperscript{5}, Stephane Proux\textsuperscript{6}, Dionicia Gamboa\textsuperscript{7}, Peter Chiodini\textsuperscript{8}, Jane Carter\textsuperscript{9}, \\
Mehul Dhorda\textsuperscript{10}, David Isaboke\textsuperscript{9}, Bernhards Ogutu\textsuperscript{11}, Wellington Oyibo\textsuperscript{13}, Elizabeth Villasis\textsuperscript{7},\\  
Kyaw Myo Tun\textsuperscript{13}, Christine Bachman\textsuperscript{1}, David Bell\textsuperscript{1}, Courosh Mehanian\textsuperscript{1} }  \\

 \IEEEauthorblockA{
\textit{\textsuperscript{1}Intellectual Ventures/Global Good Research, \textsuperscript{2}IBM (formerly IV/GGR), \textsuperscript{3}Arizona State U., \textsuperscript{4}Creative Creek LLC,}\\  
\textit{ \textsuperscript{5}LSHTM, \textsuperscript{6}SMRU, \textsuperscript{7}UPCH,  
 \textsuperscript{8}HTD,  \textsuperscript{9}Amref,  \textsuperscript{10}WWARN,  \textsuperscript{11}Kemri, \textsuperscript{12}U. Lagos, \textsuperscript{13}DSMA  } \\
Corresponding authors: \{cdelahunt, mhorning\}@intven.com 
 }
}

\maketitle

\begin{abstract}
Malaria is a life-threatening disease affecting millions. 
Microscopy-based assessment of thin blood films is a standard method to \textit{(i)} determine malaria species and \textit{(ii)} quantitate high-parasitemia infections. 
%Unfortunately,  high inter- and intra-microscopist variability in quantitation, coupled with the labor-intensive nature of manual microscopy, impacts a key use-case, viz. quantitation for anti-malarial drug resistance monitoring.
Full automation of malaria microscopy by machine learning (ML) is a challenging task because field-prepared slides vary widely in quality and presentation, and artifacts often heavily outnumber relatively rare parasites.

In this work, we describe a complete, fully-automated framework for thin film malaria analysis  that applies ML methods, including convolutional neural nets (CNNs),  trained on a large and diverse dataset of field-prepared thin blood films.  
Quantitation and species identification results are close to sufficiently accurate for the concrete needs of drug resistance monitoring and clinical use-cases on field-prepared samples.

We focus our methods and our performance metrics on the field use-case requirements.
We discuss key issues and important metrics for the application of ML methods to malaria microscopy. \\ 
\end{abstract}


\section{Introduction}

Malaria is a mosquito-borne disease caused by \textit{Plasmodium} species (\textit{P. falciparum, P. vivax, P. ovale} and \textit{P. malariae} in humans) infecting more than 200 million and killing nearly half a million people annually \cite{whoReport}.
Manual microscopy examination of Giemsa-stained blood films is a widespread malaria diagnosis method. 
Key use-cases include diagnosis; species identification (ID) to guide treatment \cite{cdcSpecies}; and quantitation of parasites for drug resistance studies, to track how fast a drug clears parasites from the blood.
However, a lack of training,  high inter-sample variability in preparation and presentation, and difficult field conditions can result in poor  accuracy \cite{zurovac,pembele}. 
Also, lack of trained personnel limits the number of drug resistance sentinel sites.

Malaria microscopy is a difficult task for automated image-processing and machine learning (ML) systems for two reasons: Field-prepared blood films vary widely in quality and presentation;  and  parasites are small (with feature size close to optical limits of resolution), rare, highly variable, and easily confused with non-parasite objects (artifacts).  
But it is also a high-value target,  due to the potential  benefit for so many people, and also because automated systems have some concrete advantages:
They can be widely deployed, solving the expert-training bottleneck;
they can examine more blood volume per patient, reducing variability in quantitation caused by Poisson statistics; 
and their results are reproducible. 
 
Thin and thick blood films have distinct uses. 
Thick  films are typically used for diagnosis and for quantitation of low-parasitemia infections, because the larger blood volume gives a lower limit of detection (LoD) and more stable parasite counts.
Thin film are used for species ID, and for quantitation of high-parasitemia infections \cite{whoMicroscopy}. 

\begin{figure*}[h!]
\begin{center}
\centerline{
\includegraphics[width=0.8\linewidth]{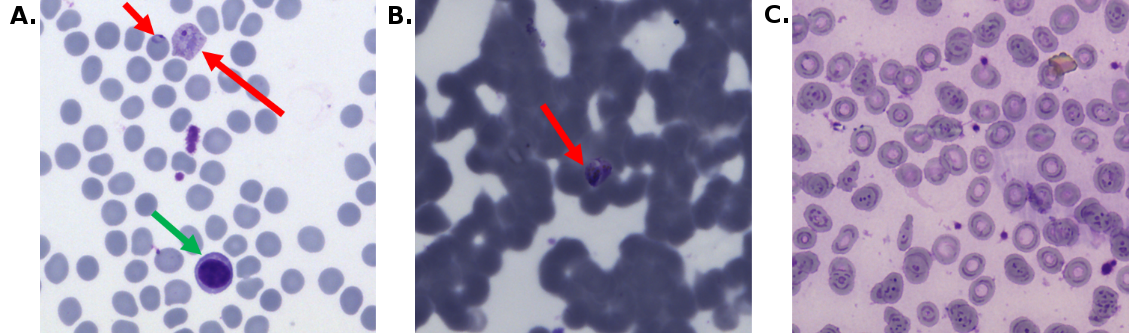}}
\caption{ Portions of typical thin film FoVs. Green arrow indicates a white blood cell; red arrows indicate parasite-infected RBCs. Other purple stained objects are distractors. (A) Ideal FoV. (B) Clumped RBCs. (C) Dirty FoV from a malaria-negative sample. }
\label{thinFilmsFig}
\end{center}
\end{figure*}

Field-prepared Giemsa-stained thin films vary greatly in presentation, e.g. in red blood cell (RBC)  color and morphology,  parasite appearance, type and number of artifacts, and amount of RBC clumping.  
 Fig. \ref{thinFilmsFig} shows typical thin film fields of view (FoVs).  

\begin{figure*}[h!]
\begin{center}
\centerline{\includegraphics[width=0.8\linewidth]{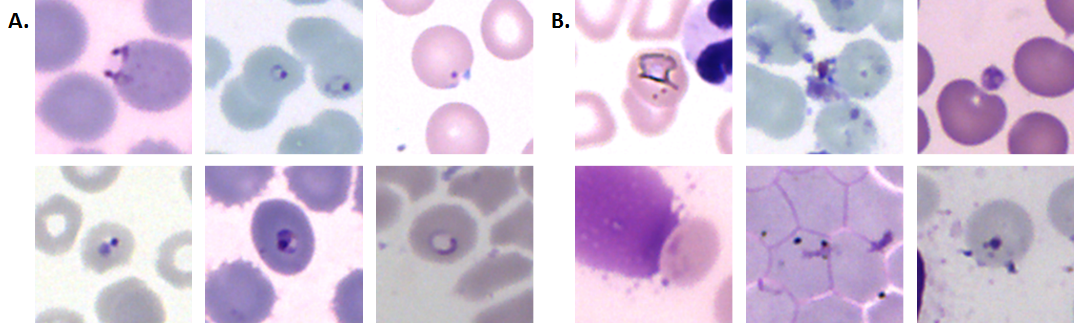}}
\caption{(A) Thin film ring-stage parasites (B) distractors.
More examples are found in S.I. \cite{suppInfo} }
\label{parasitesFig}
\end{center}
\end{figure*}

Malaria parasites display several developmental stages in blood  \cite{whoMicroscopy,garnham} starting as ring-stage trophozoites (hereafter ``rings''), seen in Fig. \ref{parasitesFig}a, then maturing into trophozoites, schizonts, and gametocytes (hereafter ``late stages''). 
Rings predominate in \textit{P. falciparum} infections, and are the targets for quantitation in drug resistance studies. 
Late stages are used for species ID.
Artifacts (``distractors'') are very common in field-prepared slides, and often heavily outnumber parasites. 
Examples are seen in Fig. \ref{parasitesFig}b and in Supplementary Information (S.I.) \cite{suppInfo}.

%-----------------------------------------------------------------------------------

A recent review \cite{poostchi} highlights several key problems with published automated malaria detection studies: 
\textit{(i)} datasets are too small;
\textit{(ii)} reported metrics are often incomplete and not comparable between studies;
\textit{(iii)} reported metrics are often object-based (not patient-based) and are thus not relevant to the clinical task, which is entirely patient-focused; and 
\textit{(iv)} train and validation sets often contain objects from the same patient.
As with other applications of ML to health care tasks, an understanding of the domain-specific constraints and use-cases is vital  but is often missing \cite{koller}. 

Studies including \cite{linder,tek1,anggraini,loddo,gopakumar,ross,abbas,le,mehanian}  discuss issues central to automated processing of thin films.
These issues include 
 \textit{(i)} the importance of sample-level variability  \cite{le, mehanian}; \textit{(ii)} the importance of false-positive (FP) rates per unit blood  \cite{tek1,mehanian} as a determinant of both LoD and quantitation accuracy; \textit{(iii)} the centrality of patient-level (not object-level) metrics \cite{linder,le,anggraini,mehanian}; and \textit{(iv)} irregularity and clumping of RBCs in thin films and the high computational cost of separating clumped RBCs \cite{linder,abbas,gopakumar}.

\begin{figure*}[h!]
\begin{center}
\centerline{\includegraphics[width=0.8\linewidth]{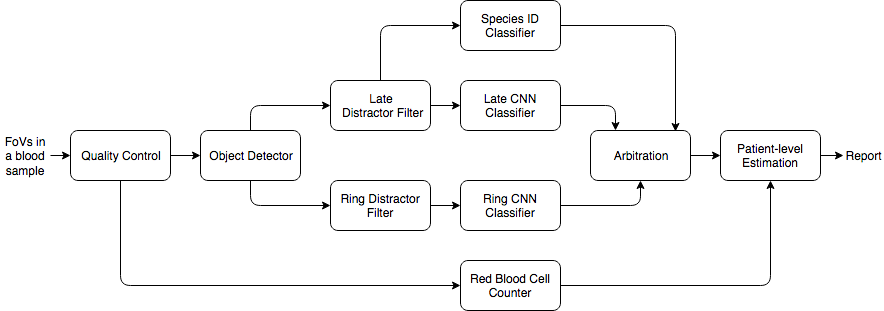}}
\caption{The modules and flow of the proposed thin film framework. 
Rings and late stages are visually distinct targets, so the two branches use different CNN architectures.}
\label{schematicFig}
\end{center}
\end{figure*}

This paper describes a complete, fully-automated thin film malaria assessment system, intended to complement the thick film system in \cite{mehanian}.
Its goals are quantitation and species ID (thick film handles diagnosis, since the greater blood volume yields a lower LoD).
The system includes two branches, one for rings and one for late stages, and modules for: FoV quality control (QC); RBC counting; object detection; distractor filters;  CNN  classifiers; species ID; branch arbitration; and patient-level disposition. 
  Fig \ref{schematicFig} gives a schematic. 

The main contributions of this work include: 
\textit{(1)} Use of field-prepared (rather than in-house) slides, sourced from clinics across four continents;
\textit{(2)} A complete, fully-automated system for patient-level results on thin blood films;  
\textit{(3)} Three classifiers arranged in series, as a means to handle the high numbers of distractors vs relatively rare parasites; 
\textit{(4)} Use of convolutional neural nets (CNNs), supplied with sufficient data for the task complexity; 
\textit{(5)} Quantitation and species ID results close to sufficiently accurate, on field-prepared slides, to meet drug resistance study and case management use-cases; 
\textit{(6)} Analysis of machines' advantage over microscopists due to reduced Poisson variability; and 
\textit{(7)} Description of patient-level metrics  that realistically target malaria microscopy use-cases, for use in algorithm evaluation.

%\section{Methods}

\section{Dataset}

A field-deployable malaria assessment system (especially one with data-hungry CNNs) requires a large and diverse set of training images, because high variability in field-prepared slides is a central challenge, and because algorithms need to generalize to as-yet unexamined slides from new clinics and regions.
In conjunction with our field partners, we assembled 798 image sets of field-prepared blood films from 765 patients, totaling 323k thin film FoV images.
Regions included South America, Africa, Asia, and London (returning travellers), encompassing over 12 countries.  
The slide collection was rich in  negative, $P. falciparum ~(Pf)$, and $P. vivax ~(Pv)$ samples, but suffered from a relative lack of $P. ovale~(Po)$ and $P. malariae ~(Pm)$ samples since these are rarer species.
We annotated over 92k objects of the four major malaria species.  
Slide metadata and object annotations were stored in a SQL database, which simplified maintenance of patient-level structure during algorithm development.

\subsection{Image capture}

Field-prepared slides were mounted with \#0 coverslips and digitized with Motic EasyScan-Go \cite{motic} automated scanning microscopes. 
The microscope has a 40x, NA=0.75 objective, infinity-corrected optical train, and 10W LED Kohler illumination. 
A CMOS camera captures images (2048$\times$1536 pixels) at an approximate pixel pitch of 8.3 pixels/$\mu$m.
A single high-quality FoV can contain 400 well-separated RBCs.
Each FoV consists of a stack of 7 z-slices, to ensure that every object has at least one in-focus version (a stack is needed to handle non-level slides and artifacts that derail the auto-focus).
All slides were prepared at field locations, and were scanned at either field locations or a central lab. 
 
\subsection{Object types}

Because parasites have visually distinct stages, we used three categories (ring, late stage, and transitional) for algorithm development. 
When stained with Giemsa, rings have one or two distinct purple nuclei and (ideally) a characteristic blue cytoplasm. 
They are typically inside RBCs but sometimes appear to be just outside (applique forms). 
We lumped all mature stages together as ``late stages.'' 
Parasites intermediate on the continuum between ring  and late stage  were labelled ``transitional.'' 
Because objects in blood films are not always identifiable, we added an important fourth annotation category, ``doubtful''.  
Excluding transitional and doubtful objects from the training set yielded classifiers with improved accuracy.

\subsection{Train/Validation sets} 

The malaria use-case  requires that objects from a given blood sample be used in either training or  validation, never in both. 
Dividing objects from a single sample can substantially improve object-level classifier results, but it is highly unrealistic.
At every point in our framework, sample-level integrity was preserved in  training-validation splits.

For training, ground-truth was defined as follows: 
Positives came from parasites (rings or late stages) annotated in our malaria database.
Parasite annotations were examined by at least two trained humans and were further vetted at least once.
Distractors were defined as any detected object that had no annotation. 
This definition of distractor required complete parasite annotations of training images, to prevent unmarked parasites being included in the distractor pool.
A way to relax  the annotation task (on validation) is described in S.I. \cite{suppInfo}.

Transitional stages and doubtful objects (i.e. objects that may or may not have been parasites) were excluded from both positive and negative pools.
This allowed the classifiers to focus on a single parasite type (ring or late stage).
Transitional parasites, though excluded from training, were still  accurately identified as parasites by one or other of the branches.

\section{Algorithm framework}

\subsection{Constraints on architecture}  
Algorithm architecture was constrained by the need for low computational complexity.
On average, each blood slide generates $\sim$200 FoVs, each a stack of 7 images.
Each FoV may yield multiple objects to process.
Processing time was limited to $\sim$15 minutes on a standard laptop CPU (no GPU). 
Since parasites are rare and greatly outnumbered by distractors, the object detector and classifiers must have high sensitivity (percentage of parasites correctly classified) and also very high specificity, i.e. a low False Positive (FP) rate.
 
Generic deep learning based detection methods  \cite{liu,redmon,ren,he} have achieved strong results on natural images. 
However, two-stage detectors, such as SPPnet \cite{he} and Faster R-CNN  \cite{ren} are too slow for this application, while faster  
single-stage detectors, such as SSD  \cite{liu} and YoLo  \cite{redmon} have lower performance. 
Both single- and two-stage detectors have unacceptably poor accuracy on small objects.
To combine fast processing and sufficient object-level sensitivity and specificity, we used three classifiers in series: an initial detector and  a distractor filter using low-cost manual features, and finally a CNN.

\subsection{System overview}
Given an FoV, the system first runs quality control, counts RBCs, and runs initial detection of objects.
It then splits into two branches, one for rings (for quantitation), and one for late stages (for species ID).
Each branch has two parts:
First, a high-sensitivity distractor filter culls  the bulk of the more obvious distractors; then a CNN classifies the remaining objects as parasites or distractors.
The detected late stage parasites also pass through a species ID module.
Finally, the various outputs are combined to deliver patient-level quantitation and species ID predictions.

The following subsections describe the various modules.

\subsection{Quality control module}  
Quality of slide preparation and image acquisition varies widely in the field. 
The QC module identifies and culls  FoVs that are blurry or empty. %, or that contain  heavily-clumped RBCs. 
A FoV may be blurry when the thickness of the cover slip on the slide is not compatible with the scanning microscope, the microscope is unable to focus on the slide, or when it focuses on objects on a higher plane than the blood film.   
%Depending on the layers of RBCs and stain, the color of the RBCs vary. 
FoVs  that are empty or blurry tend to have  pixel values within narrow ranges. 

Despite proposed image quality assessment methods \cite{pertuz,kang,jaiswal}, blur detection remains challenging because image content can affect image sharpness. 
Further, our application requires a no-reference, fast method.
We apply a two-stage QC process.
First, we calculate the standard deviation (std dev) of grayscale pixel values and the dynamic range of the gradient of the grayscale FoV. 
If either of these is lower than pre-set thresholds, the FoV is rejected.
Second,  we calculate focus metrics on the FoV and on a corresponding artificially-blurred version \cite{jaiswal}. 
Using the method in \cite{jaiswal} we selected a small, computationally-efficient subset of these focus metrics as features for blur detection, then trained gradient boosted trees (GBT) \cite{chen} classifier to identify blurry FoVs. 

\subsection{RBC counter module}  
Accurate RBC counts are needed for quantitation of \textit{Pf} rings. (Species ID depends on distinguishing different parasite morphologies, not RBC count.)
Only high parasitemia samples are quantitated on thin film because at low parasitemias, the lower blood volume causes high Poisson  variance in ring counts (low densities are quantitated on thick film). 
We can thus assume that parasitemia is high: for microscopists, over 5k p/$\mu$L or 16k p/$\mu$L depending on protocol \cite{whoMicroscopy}; 
for automated systems (e.g. \cite{mehanian}), over 80k p/$\mu$L. 

We estimate  ring branch parasitemia per $\mu$L = $\hat{P}$ as follows (late stage is similar):
 \begin{equation}
 \hat{P}= \frac{nR(\frac{5e6}{nRbc}) - \hat{fp}}{\hat{s}}, \text{ where}
 \label{estQuantEqn}
 \end{equation}
$~~~$$nR$ = number of alleged rings found by the algorithm, i.e. the number of ring branch objects with CNN scores above some threshold $tRing$, \\
$~~~~$$\hat{fp}$ = expected number of FPs/$\mu$L, \\
$~~~~$$\hat{s}$ = expected sensitivity of the ring classifier, \\
$~~~~$5e6 = number of RBCs/$\mu$L, and \\
$~~~~$$nRbc$ = number of RBCs counted.\\
Hyperparameters such as $tRing, \hat{s}, \text{ and } \hat{fp}$ are determined on a validation set ($\hat{fp}$ uses negative samples only).
$\hat{s}$ could be the mean (or median) of sample  sensitivities  over all positive samples $\mu(\textbf{\textit{s}})$, and $\hat{fp}$ could be the mean (or median) of sample FP rate  over all negative samples $\mu(\textbf{\textit{fp}})$. 
 
Error in the RBC count $nRbc$ directly impacts quantitation accuracy through the denominator of the second term of Eqn.\ref{estQuantEqn}, and must be minimized.
	                      
The biggest challenge in RBC counting is to accurately segment  clumps of overlapped RBCs, which are common in field-prepared thin films. 
However,  automated scanning microscopes can image more FoVs than needed, which  enables the following strategy: 
Ignore clumped RBCs and use only single (and double) RBCs, which are easily counted \cite{linder}. 
We count the single RBCs as we scan, directing the microscope to continue collecting FoVs until 20k single RBCs have been tallied (a human microscopist examines 5k RBCs). 
This is generally possible even on field slides, and suffices to mitigate Poisson variance error given high parasitemia (at 80k p/$\mu$L, 20k RBCs yield roughly 320 parasites). % , mainly in a narrow band just inside the border of the thin film region. 
 
In a given FoV, we detect RBCs with simple binary gray-scale clustering, and divide the detected RBCs into singles and clumps based on blob size. 
We count the single RBCs, and create a ‘quantitation mask’ containing only these RBCs plus a margin to capture applique rings.  
This RBC count is highly accurate (i.e. $nRbc$ in Eqn.\ref{estQuantEqn} has very low error). 
Only rings within the quantitation mask are included in $nR$ in Eqn.\ref{estQuantEqn}. 

We also detect and classify all objects in the FoV. 
For species ID, we use all the suspected late stages, whether in single or clumped RBCs, since only their morphology matters.   

\subsection{Object detector module} 
The object detection module generates a list of candidate objects in each FoV.
A domain-specific detail enables a simple yet effective object-of-interest detector. 
Giemsa  stain colors DNA (e.g. parasite nuclei, WBCs) purple and RNA (e.g. in cytoplasm) blue, modulo variations due to pH. 
RBCs and background stain to pink, green, or gray.  
 
We project the color image to gray scale  via 
\begin{equation}
gray = \frac{RB}{G^2 + \epsilon}
\label{grayscaleEqn}
\end{equation}
where R, G and B are the red, green and blue channels of the color image. 
This grayscale image highlights purple pixels and suppresses green pixels in the image, 
and detects most parasite candidates (as well as distractors that stain purple). 
%Unlike \cite{mehanian} this method is independent of sample pixel statistics.   
The grayscale image is thresholded pixel-wise using dynamic local thresholding \cite{mehanian}. 
Candidate objects are chosen by finding connected-components in the thresholded image. 
%The candidate objects are separated into ring-stage and late stage candidates using area of the object. 
%Late stage objects are generally larger than ring-stage objects.  
%To accommodate the similar sizes of rings and trophozoites, there is a small overlap between ring and late stage objects.

This method is applied to all z-slices in a given FoV. 
Distance-based clustering groups together instances of the same object detected in multiple z-slices, to account for microscope stage jitter and variation in x, y coordinates due to focus. 
The most in-focus object, i.e. the object with the highest Brenner focus score \cite{brenner}, is retained for further processing. 

\subsection{Distractor filter module(s)}  
Malaria parasites are rare in thin blood film images (1 per 100 RBCs in high parasitemia cases, and more commonly fewer than 1 per 1000 RBCs), so distractors typically outnumber parasites. 
The detector needs high sensitivity, high specificity, and computational efficiency.  

Many detected distractors can be efficiently culled via manual features. 
We trained GBT  classifiers, one for each branch, using region properties of the detected objects as features (see S.I. for details).
%Properties include area, intensity, extent, Euler number, eccentricity, mean gradient, prominence, bumpiness, roundness, ridgeness, donut shapeness, external contrast, internal contrast, mean of red, green and blue channels, coefficient of variation of gray scale, gradient, red, green and blue channels.
The distractor filters achieved 0.96 (ring) and 0.94 (late stage) areas under the  ROC curve on a validation set,
%The FROC  is similar to the ROC but plots x-axis = FP/$\mu$L, to account for imbalance between classes.
and culled most potential distractors. 

\subsection{CNN classfier module(s)}  
In the second stage of each branch, a CNN classifier distinguishes parasites from the remaining, most difficult distractors. 
CNNs are state-of-the-art technology in many computer vision  \cite{lecun}  and biomedical image processing  \cite{litjens}  applications. 
Published CNN architectures are most often designed for large-scale datasets (e.g. ImageNet \cite{russakovsky} with 1000 output classes), so they overfit  our dataset. 
We therefore tailored CNN architectures for  our domain-specific case, with two output classes (parasite vs distractor). 
We explored various architectures, including Inception-style networks   \cite{szegedy}, fully convolutional networks  \cite{springenberg}  and VGG-style networks   \cite{simonyan}.
We developed and tested CNNs in Caffe  \cite{jia} using  cross entropy loss with stochastic gradient descent. 
 
The ring branch CNN had 3 convolutional layers, followed by two Inception modules and one fully connected layer. 
This architecture enabled  identification of features at multiple scales, e.g. small features near the nucleus and larger features in the cytoplasm of the parasite.
The Inception modules had convolutional layers with kernel sizes 1$\times$1, 3$\times$3, and 5$\times$5  and a dimensionality reduction kernel of size 1$\times$1. 
The number of kernels within each branch of the Inception module was chosen such that the number of parameters to be learned in each Inception module was equal. 
Thus not all multi-scale features were weighted the same. 
Deeper layers had more filters per convolutional layer. 
Thumbnails were 64$\times$64 pixels. 
 
The late stage branch CNN was a fully convolutional neural network with  7 convolutional layers and increasing number of kernels as network depth increased.  
Spatial reduction was achieved by strided convolutions which could learn the spatial reduction operation.
Thumbnails were 144$\times$144 pixels. 

The thumbnails had 4 channels, namely, red, blue, green and an inverse gray channel. 
The 4\textsuperscript{th} channel gave stronger test accuracy. 
Thumbnails were augmented in three ways: Flipping and rotating (90 degree increments); 
random horizontal or vertical spatial translations of  pixels; 
and random gamma transformation of each color channel as in  \cite{mehanian}. 

Weights were initialized by the Xavier method \cite{glorot}. 
Other parameters included: Weight decay 5e-4, momentum 0.9, batch size 128, ``poly'' learning policy with learning rate 1e-3, dropout 0.3 (late branch) and 0.5 (ring branch).
CNN architectures are shown in S.I. \cite{suppInfo}. %\\ \\  
% \noindent \textit{Training object restrictions:}
 
Distractors typically outnumber total parasites, because  distractors are derived from all (not just positive) samples,   parasites are relatively rare objects, and field slides can be distractor-rich.
The number of distractors selected from each training sample was capped, both to keep training imbalances within 2- or 3-to-1,  and to ensure that  a few very dirty samples did not dominate training. 
To get a training set that covered the whole distribution of distractors, while emphasizing difficult types, we randomly selected 80\%  distractors from those that passed the distractor filter (i.e. from the relatively difficult distractors), and the remaining 20\%  from the remainder. 

To avoid the CNN training set being dominated by a few high parasitemia samples, the number of parasites each sample could contribute to training was capped.
When assessing network training, we watched for sample-level effects. 
%For example, if one validation sample had many very faint parasites, it could distort apparent object-level accuracy, since all parasites in the sample responded similarly; but in CNN terms, the failure was on one parasite presentation only. 
For example, if one high parasitemia validation sample had faintly-stained parasites that largely went undetected, it would disproportionally affect the object-level statistics. 
However, it would only represent one failure mode in the CNN, viz failure to detect faint parasites.

\subsection{Species ID module}  
The four malaria species have very similar ring forms, while the mature (late) stages exhibit distinctive features. 
Thus the species ID module tries to identify species of objects detected and classified as parasites by the late stage branch. 
Geographic priors are not used (though these can be very informative \cite{autino}).
Due to the many species and variety of late stage forms, the classifier has 13 categories:   
Four ring categories, i.e. one for each species (\textit{Pf, Pv, Po, Pm}); similarly four transitional and four late stage categories; plus distractors. % as the 13\textsuperscript{th} category.

We trained a 13-class GBT classifier that used manual features (details in S.I.) on the segmented late stage objects in each thumbnail.
Each training sample was allowed to contribute a maximum of 100 objects per output class, to ensure wide sample-level variety. 
%As with the CNNs and distractor filters, the training and validation sets were divided by sample.
To aid segmentation of objects, we enhanced contrast as follows: We converted the thumbnail to the luminance and chrominance space; performed adaptive histogram equalization so that the pixels values followed a Raleigh distribution; converted back to RGB color space; and morphologically eroded using a ‘ball’ structuring element. 
Foreground and background were found via k-means using the luminance and chrominance of this enhanced image, and the greyscale image (Eqn.\ref{grayscaleEqn}), as features.
 
%Any holes in the foreground image are filled and features are extracted on this sub-image. 
%The manual features included standard region properties such as elongation, perimeter, major and minor axis length, histogram of the grayscale image, pixel statistics of the gray-scale image; and properties of the gray-level co-occurrence matrix \cite{haralick} such as contrast, correlation, energy and homogeneity.

We also built a CNN for species ID, using transfer learning \cite{yosinski} due to the smaller numbers of late stage parasites for training.
Calendar constraints prevented us from testing it. 
%We suspect a CNN would work as well or better than the method above.  
 
The parasite stage classification probabilities are used by the patient-level disposition module to predict malaria species.% for the patient.

\subsection{Object Arbitration module}  
Because the ring and late stage branches of the decision tree architecture each apply their own distractor filters to the set of detected objects, there will be three kinds of objects: 
\textit{(1)} detected only by the ring branch; \textit{(2)} detected only by the late branch; and \textit{(3)} detected by both branches. 
The arbitration module decides, for objects detected by both branches, whether they should be judged as possible rings or as possible late stages.
%This yields a single list of objects. % along with their associated species/stage probabilities. 
%The output takes the form of a table that lists 13 probabilities for each object, one row per object. 
%The thirteen species/stage categories for which probabilities are computed are listed in Table 1.
Modulo complications (not discussed) due to the species ID module, an object's proper category is decided simply by which branch gave it a higher CNN score.

\subsection{Patient-level disposition module} 

\subsubsection{Quantitation}
Estimated parasitemia is the sum of ring and late stage parasitemias, each made according to Eqn. \ref{estQuantEqn}.
Drug resistance studies require only quantitation of \textit{Pf} rings. 

\subsubsection{Species ID}
Species ID primarily uses late stage forms. 
But it also considers ring counts (sometimes from thick film) because of two unique \textit{Pf} traits: % that necessitate yet enable identifying it using only immature ring forms: 
mature \textit{Pf} parasites sequester in the microvasculature, so \textit{Pf} typically presents only ring parasites in blood samples \cite{garnham};
 and \textit{Pf} can reach much higher ring densities than other species.
 
Thus, predicting the species of malaria parasites infecting a patient is based on three factors: the species probabilities of the late stage parasites;  the ring density per $\mu$L;  and the ratio of ring and late stage parasite counts. 
 
For the late stage species prediction, we sum the species probabilities of all late stage parasites (i.e. objects with late branch scores above a threshold $tLate$). 
This sum up-weights objects with strong likelihood of one species and down-weights objects with uncertain species.
The highest sum gives the predicted species.
%As described above, detected objects receive CNN scores from both the ring and late branches (which may be zero if undetected in either branch), and, when detected in the late branch, probabilities of belonging to each of the 13 species categories listed in table 3. 
%Three criteria are applied when deciding which objects to identify as late stage for predicting species: 
%The object must be arbitrated as late stage based on CNN scores, the late stage CNN score must be above a threshold, and the ‘distractor’ species probability must be below a different threshold.  
%From the table of objects that meet these criteria, we use a simple heuristic for predicting the species: 
%The probabilities of each class over all the objects are summed (resulting in a 13x1 vector), then the “interim-stage” and late stage scores for each species are summed, ignoring the distractor and ring classes (resulting in a 4x1 vector corresponding to the four species).  
%The late stage species is then predicted as the class with the maximum summed probability. 
 
%The ring-to-late stage ratio is used as follows:
If either the density of ring forms or the ratio of rings to late stages are above empirically determined thresholds, the species is reported as \textit{Pf}.  
Otherwise, the late stage species prediction is reported. % (which may be \textit{Pf} or another species).  
An exception occurs when there is both a high density of parasites, indicating the presence of \textit{Pf}, but also a  high number of late stages (atypical for \textit{Pf}).  
Then a mixed infection is reported: \textit{Pf} and the late stage species prediction.

The algorithm delivers a patient report with parasitemia, species, and thumbnails of top-scoring objects for use by technicians.
Typical reports are shown in S.I. \cite{suppInfo}.

\section{Results}

\subsection{Important metrics} \label{metrics} 

Patient-level results are by far the most relevant to the malaria use-case, for diagnosis, quantitation, and species ID.
This section describes key Figures of Merit (FoMs)  which guided our development and assessment of algorithms.
%Limit-of-detection FoMs for diagnosis (relevant to thick films) can be found in \cite{mehanian}.

\subsubsection{Quantitation error}
Three forms of error affect quantitation: \textit{(i)} RBC counts; \textit{(ii)} parasite counts; and \textit{(iii)} irreducible Poisson error.

RBC counting errors contribute  to quantitation error in a straightforward way via the second term of Eqn.\ref{estQuantEqn}, $\frac{5e6}{nRbc}$. 

Parasite counting errors  (first term of Eqn.\ref{estQuantEqn}) stem from sample-level variations in sensitivity  and in FP rate (derivation is given below).
Thus, two FoMs for quantitation error are
\begin{equation}
{\frac{\sigma(\textbf{\textit{s}})}{\mu (\textbf{\textit{s}})} } \text{ and } {\frac{\sigma(\textbf{\textit{fp}})}{\mu (\textbf{\textit{s}}) } \frac{1}{P} } \text{, where} ~~~~~~~~ 
\label{quantFomsEqn}
\end{equation}
$~$$\sigma(\textbf{\textit{s}})$ = std dev of sample  sensitivities (over all samples),\\
$~~$$\mu(\textbf{\textit{s}})$ = mean of sample  sensitivities (over all samples), \\ 
$~~$$\sigma(\textbf{\textit{fp}})$ = std dev of sample  FP rates per $\mu$L (over all samples).\\
$~~$$P$ = parasitemia per $\mu$L.\\
At high parasitemias (i.e. the thin film use-case) the first term dominates because the second term shrinks as $1/P$.
 
Irreducible Poisson variation affects the actual number of parasites contained in the examined blood, and will result in different counts if a perfect counter examines two distinct sections of a film.
The magnitude of this variation depends on parasitemia and number of RBCs examined.
It can be mitigated by high parasitemias and by examining high numbers of RBCs.
Automated systems have a powerful advantage over microscopists in this regard. 
See S.I.\cite{suppInfo} for discussion. 

\subsubsection{Derivation of Eqn.\ref{quantFomsEqn}}
Eqn.\ref{estQuantEqn} gives the estimated parasitemia $\hat{P}$ of a patient $p$. 
The first term contains error from multiple sources.
For a  patient $p$, let $tp$ = number of TPs found, $fp$ = number of FPs found, $s$ = sensitivity on this sample, with other terms defined as in Eqn.\ref{estQuantEqn}.
Then the first term 
 \begin{equation}
 \frac{nR - \hat{fp} }{\hat{s}} = \frac{tp  + fp  - \hat{fp} }{\hat{s}}
 \label{sourceQuantErrorEqn}
 \end{equation} 
 Let $\Delta tp = tp  - P\hat{s}$ , i.e. the discrepancy between our actual TP count and the count we would get if $s  = \hat{s}$.\\ 
 Let $\Delta fp = fp  - \hat{fp} $, i.e. the discrepancy between our actual FP count and the count we would get if $fp  = \hat{fp}$.
 Then 
 \begin{equation}
    \frac{tp + fp - \hat{fp} }{\hat{s}} = \frac{(P\hat{s} + \Delta tp) + (\hat{fp} + \Delta fp) - \hat{fp} }{\hat{s} }
    \label{deriv1Eqn}
 \end{equation}
 \begin{equation}
    = {\frac{P\hat{s}}{\hat{s} }{( 1 + {\frac{\Delta tp}{P\hat{s}} } + \frac{\Delta fp}{P\hat{s}} } ) }.
    \label{deriv2Eqn}
 \end{equation}
 The leading term of Eqn. \ref{deriv2Eqn} is the true parasitemia $P$.
 So the relative error $relErr$ on patient $p$ is
 \begin{equation}
 relErr(p) = {\frac{\Delta tp}{P\hat{s}} } + {\frac{\Delta fp}{P\hat{s}} } = {\frac{(s - \hat{s})P}{P\hat{s}} } + {\frac{\Delta fp}{P\hat{s}} } 
 \end{equation}
 \begin{equation}
 = {\frac{\Delta s}{\hat{s} } } + {\frac{\Delta fp}{\hat{s} P } }  \text{, where } \Delta s = s - \hat{s} \text{, i.e. the discrepancy}
 \label{individualErrorEqn}
 \end{equation}  between actual sensitivity and expected sensitivity. 
 
This implies that for the population of samples, parasite counting error can be characterized by
 \begin{equation}
  {\frac{\sigma(\textbf{\textit{s}})}{\mu (\textbf{\textit{s}})} } + {\frac{\sigma(\textbf{\textit{fp}})}{\mu (\textbf{\textit{s}}) } \frac{1}{P} } \text{ where }
 \label{finalLineOfDerivEqn}
 \end{equation}
$~~~~$ $\mu(\textbf{\textit{s}})$ = mean of sample sensitivities (standing in for $\hat{s}$),  \\
$~~~~~$ $\sigma (\textbf{\textit{fp}})$ = std dev of sample FP rates  (over all samples).
 
 The  two terms of Eqn. \ref{finalLineOfDerivEqn} are the FoMs given in Eqn.\ref{quantFomsEqn}.
 The first term is the error due to variance (over all samples) of sample sensitivities, scaled by mean sample sensitivity. 
 This error can be reduced by increasing overall sensitivity and/or by reducing variation in sensitivity by sample.
 
 The second term is the error due to variation in FP rates.
 It decreases as $\frac{1}{P}$, so it is a dominant effect at low parasitemias (it is a noise floor in diagnosis and LoD calculations) but a minor effect at high parasitemias.% (including the thin film quantitation use-case).
 
 As discussed in \cite{tek1}, we can trade off sensitivity and specificity at the object level according to our goal, by varying threshold operating points. 
 For diagnosis, very low $\sigma(\textbf{\textit{fp}})$ in Eqn. \ref{finalLineOfDerivEqn} is needed to achieve low LoD. 
 On the other hand, to quantitate high parasitemia samples one must minimize $\frac{\sigma(\textbf{\textit{s}})}{\mu(\textbf{\textit{s}})}$ %\\
 %$~$\\
 in Eqn. \ref{finalLineOfDerivEqn}, while $\sigma(\textbf{\textit{fp}})$   can be larger. 
 Operating points with low FP rates typically have lower sensitivities (lower left of the ROC curve). 
 Since the two terms of Eqn.\ref{finalLineOfDerivEqn} move in opposite directions,  operating point depends on the goal.% (diagnosis or quantitation). 
 
 %$~$\\
 %$~~~$We  note that the mean (or median) sample FP rate
 We  note that the mean (or median) sample FP rate $\mu(\textbf{\textit{fp}})$ does not affect error: Expected FP rate $\hat{fp}$, which can be set to  $\mu(\textbf{\textit{fp}})$, is subtracted out (see Eqns \ref{estQuantEqn},\ref{sourceQuantErrorEqn}, and \ref{deriv1Eqn}).
 $\mu(\textbf{\textit{fp}})$ does loosely correlates with $\sigma(\textbf{\textit{fp}})$, which is a relevant quantity.

% NOT NEEDED:
%\begin{equation}
%errQ(p) = {\frac{\Delta s(p)}{\hat{s} }  }+ {\frac{\Delta FP(p)}{\hat{s}P(p)} } + \text{Poiss}(P(p),nRbc)  \text{, where} 
%\label{quantErrorsEqn}
%\end{equation}
%$~~~~$$\hat{s}$ = expected sample-level sensitivity, \\
%$~~~~$$\Delta s(p)$ = difference between $\hat{s}$ and the actual sensitivity on patient $p$,  \\
%$~~~~$$FP(p)$ = FPs/$\mu$L on patient $p$, \\
%$~~~~$$P(p)$ = parasitemia of patient $p$, and \\
%$~~~~$$Poiss( )$ is irreducible Poisson variation (which depends on parasitemia and number of RBCs examined). 

\subsection{Comparison to other methods}  
It is customary to provide a comparison of results for a proposed method vs other methods in the literature.
This is problematic here because, as discussed in  \cite{poostchi} most studies do not give patient-level results, due to data limitations and/or chosen methodologies. 
For example, if a method used a train/val split that allows a sample's objects into both train and validation, or if it did not report patient-level results, then its results are not comparable to ours.

Due to this lack of common metrics, we do not provide a comparison table.
This is not to cast shade on prior work:  
The lack of comparability is in large part due to our good fortune in having a large, varied dataset.
We can offer some comparisons to certain prior results from the literature, %for example the excellent \cite{linder}.
%We note an important 
with a caveat that the prior studies listed here used clean  in-house slides and counts, while our results used field-prepared slides, which vary more widely and often contain  more distractor objects.

\subsubsection{Median quantitation error}
Linder et al. \cite{linder} attain 21\% median quantitation error on 17 slides (20k - 40k RBCs per sample).
Le et al. \cite{le} report 20\% median  quantitation error (but with very small samples, $\approx$1800 RBCs per sample).
Our method had 18\% median  quantitation error vs in-house counts on 24 holdout slides (20k - 80k RBCs per sample), and 31\% median error vs field counts on 81 holdout slides (see Fig. \ref{quantPlotFig}).

\subsubsection{FP rates}
By Eqn. \ref{quantFomsEqn} the key FoM for FP rate is $\sigma(\textbf{\textit{fp}})$.
However, previous studies do not report this, and we can only calculate  a (very) rough proxy, namely  $\mu(\textbf{\textit{fp}})$, as follows (ring case): 
5k (Linder \cite{linder}), 12k (Tek \cite{tek1}), 15k (Anggraini \cite{anggraini}), 25k (Ross \cite{ross}), 70k (Gopakumar \cite{gopakumar}).
Our method, set to a diagnosis operating point, has $\mu(\textbf{\textit{fp}})$ = 1.6k.
This operating point gives 90\% sample-level specificity on holdout sets (i.e. 90\% of negative samples are correctly diagnosed as negatives), a requirement based on the ``WHO 56'' evaluation method \cite{whoMicroscopyQA}.

\subsection{Quantitation results}
We report results for \textit{Pf} rings since these are the most important quantitation target, as used in drug resistance studies, where error should ideally be under 25\%   \cite{whoMicroscopyQA,dhorda,white}.
Rings are a more difficult target than late stages, due to their small size and similarity to distractors.
 
\subsubsection{Object-level results} 
Object-level results are relevant only as an interim step to patient-level results.  
Also, object-level specificity and area under ROC curve (AUC) depend on the raw number of distractors but do not reflect their difficulty, so these metrics can be boosted arbitrarily by surplus easy distractors.
Considering the distractors that passed both object detection and the distractor filter,  the CNNs had 0.99 AUC (both ring and late stage CNNs), while  validation accuracy was 94.8\% (ring) and 96.6\% (late). 

\subsubsection{Patient-level results} 
In the ring branch, $\frac{\sigma(\textbf{\textit{s}})}{\mu(\textbf{\textit{s}})} = 0.13$, and $\frac{\sigma(\textbf{\textit{fp}})}{\mu(\textbf{\textit{s}})} = 6000$ for P $>$ 60k/µL. 
Based on Eqn. \ref{finalLineOfDerivEqn}, we expect %\\
%$~$\\\\
our ring quantitation error to usually be less than \\ $(0.13 +  \frac{6000} {P}) < (0.13 + 0.1) = 23\%$.

\begin{figure}[htbp]
\centerline{\includegraphics[width=0.49\textwidth] {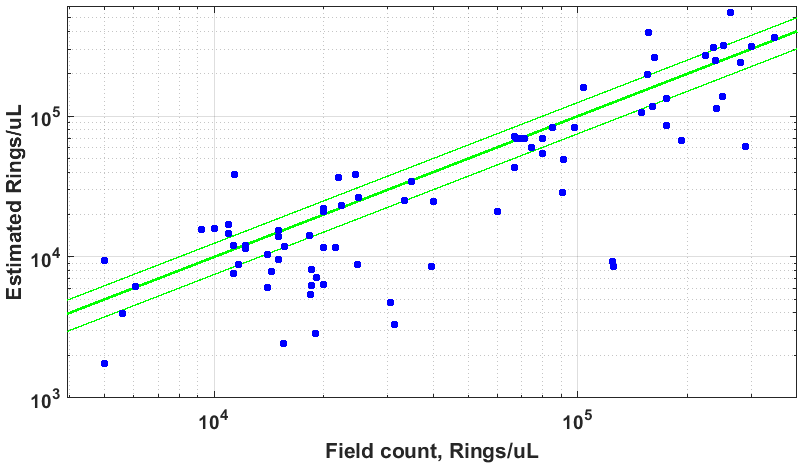}}
\caption{Quantitation accuracy on 81 holdout samples: Proposed method counts ($y$-axis) vs field counts ($x$-axis), with green +/- 25\% error lines. }
\label{quantPlotFig}
\end{figure}

Estimated quantitations, on a holdout set of 81 \textit{Pf} from 10 clinics, are shown in Fig. \ref{quantPlotFig}. 
As noted in \cite{linder}, in-house parasitemia counts are preferable for use as ground truth since  field counts are highly variable due to Poisson variability and the difficulty of manual RBC counting. 
Also, quantitations on the thick vs thin film can differ by 30\%  due to wash-off and different methods used on the two types of films.
Here we compare to field counts  to allow a larger holdout set.
38\% of holdout samples had under 25\% discrepancy vs field counts, and 50\% of the holdout set had under 33\% discrepancy. 

Algorithm undercounts are typically due to poor sensitivity, e.g. on a slide with faintly-stained parasites.
Overcounts are typically due to high FP rates. % , e.g. on dirty slides with many convincing distractors.
Discrepancies may also be due to Poisson variability \cite{suppInfo} and errors in the field counts.

\subsection{Species ID results}

WHO's 56-slide evaluation requires 90\% accuracy for expert level \cite{whoMicroscopyQA}.
Our algorithm attains this accuracy on \textit{Pf} and \textit{Pv}, but not \textit{Po} and \textit{Pm} (perhaps due to less training data).
Table \ref{speciesTable} shows our algorithm's species ID results on 42 holdout samples (10 $Pf$, 20 $Pv$, 9 $Po$, 3 $Pm$).
Since the thick film

\begin{table}[htbp]
 \caption{    \normalfont{ Confusion matrix for species ID predictions. 
    Rows are true species, columns are predictions.   
    Red  indicates errors that would affect treatment.  
    $Pf$ values in parentheses assume  tandem use with the thick film method \cite{mehanian}. $~~~~~~~~~~~ $     }     } 
  \begin{center} 
\label{speciesTable}
    \begin{tabular}{c|c|c|c|c|c|c} % <-- Alignments: 1st column left, 2nd middle and 3rd right, with vertical lines in between
                 & \textbf{Pf}  & \textbf{Pv} & \textbf{Po} & \textbf{Pm} & \textbf{\% correct}    \\
      %$\alpha$ & $\beta$ & $\gamma$ \\
      \hline
      \textbf{Pf}  & 7  &   {\color{red}2}  &  {\color{red}1}  &  {\color{red}0} &  70 (94)    \\
      \textbf{Pv} & {\color{red}1} & 28  &   1  &  {\color{red}0}   & 93    \\
      \textbf{Po} &  {\color{red}2}   &  3  &   4  &  {\color{red}0}  &  44     \\
      \textbf{Pm} &  0   & { \color{red}1}  &     {\color{red}0}  &  2  &  67    \\
    \end{tabular}
  \end{center}
\end{table}

\noindent  method in \cite{mehanian} can diagnose $Pf$ very accurately, our thin film algorithm defers to a thick film ``$Pf$'' prediction.
Important species ID errors include mislabeling $Pf$ as some other species, and mislabeling $Pv$ or $Po$ as $Pf$ or $Pm$, since these can lead to incorrect treatments \cite{cdcSpecies}.

\section{Discussion}

Malaria assessment using microscopy blood films is a difficult but high-value target for machine learning.  
The fully-automated thin film system presented here delivers accuracy that is close to sufficient for quantitation and species ID use-cases in the field.
Crucially, it works with field-prepared slides.

We have found that  to produce clinically useful algorithms, one must focus on the particular needs of the malaria use-case, including \textit{(i)} patient-level deliverables, \textit{(ii)} computational contraints, and \textit{(iii)} the high variability of field slides. 

Use-case deliverables requires metrics focused on the patient-level, since standard ML metrics such as ROC curves are insufficient to assess patient-level accuracy.

Computational constraints  sometimes require forgoing certain methods (e.g. R-CNN, Hough circle detection) and finding simpler, faster approaches.

High slide variability requires sufficient variety and quantity of training slides from many clinics to capture patient-level variation;   training and validation sets organized at the patient (not object) level;  pre-processing methods to normalize images, and classifiers robust to variations in slide presentation; and methods to minimize inter-sample variance in parasite sensitivity and FP rates, since these are the main sources of error in quantitation and diagnosis. 
% \textit{(iii)}
%On the other hand, algorithms can be brittle and vulnerable to new, unforeseen conditions in the field.
%Trained human microscopists are much more adaptable to new conditions \cite{torres}.
%Also, hardware is subject to an array of mechanical failure modes, especially in remote settings with less-developed infrastructure.
  
Conversely, one can leverage domain-specific details to simplify the task.
Examples include using the particular effects of Giemsa staining, shortcuts to RBC counting, and the assumption of high parasitemias during thin film quantitation.
Also, machines have some intrinsic advantages over human microscopists (to offset their nontrivial drawbacks \cite{torres}),
 including reduced Poisson variance and lower RBC counting error.
Also, algorithms do not fatigue, their results are replicable, and new units need no extra training. 
%: 
%Algorithms can count RBCs more accurately, which reduces quantitation error. 
%Algorithms can examine more RBCs, which reduces Poisson variance error. 

Honoring the constraints imposed, and leveraging the advantages offered, by the use-case requires close consultation with experts working in the field.
In our experience, their domain expertise is of first importance when developing  algorithms.\\ \\ \\ \\

\section{Supplementary Information}

\subsection{Malaria species and stages} 
\begin{figure*}[h!]
\begin{center}
\centerline{
\includegraphics[width=0.75\linewidth]{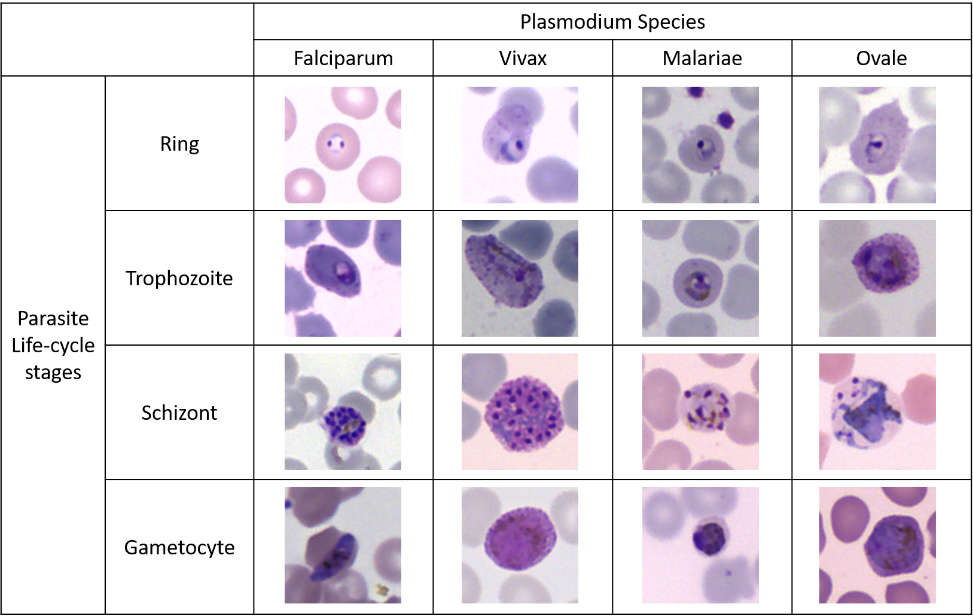}}
\caption{Examples of the four $Plasmodium$ species at different developmental stages.}
\label{allStages}
\end{center}
\end{figure*}

\begin{figure*}[h!]
\begin{center}
\centerline{
\includegraphics[width=0.8\linewidth]{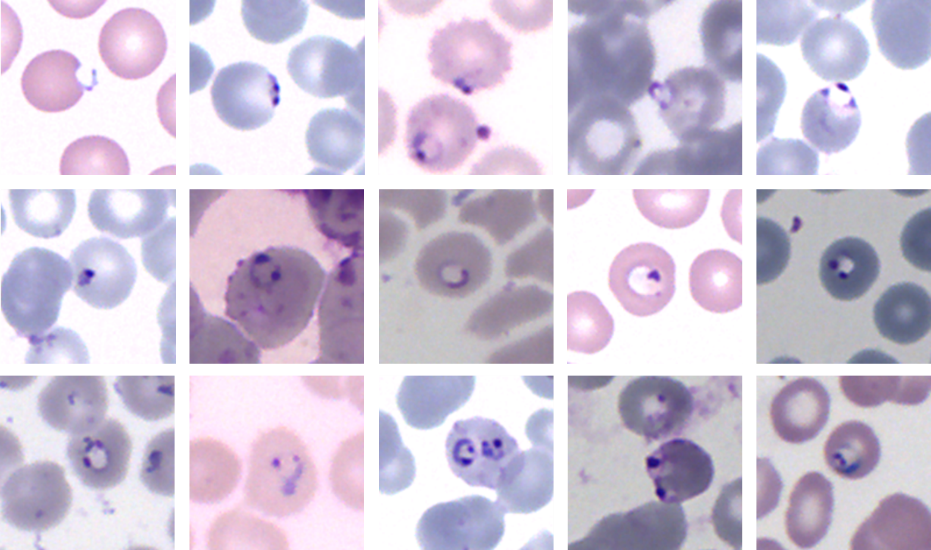}}
\caption{Ring-stage malaria parasites from field-prepared thin blood films.}
\label{rings}
\end{center}
\end{figure*}

\begin{figure*}[h!]
\begin{center}
\centerline{
\includegraphics[width=0.8\linewidth]{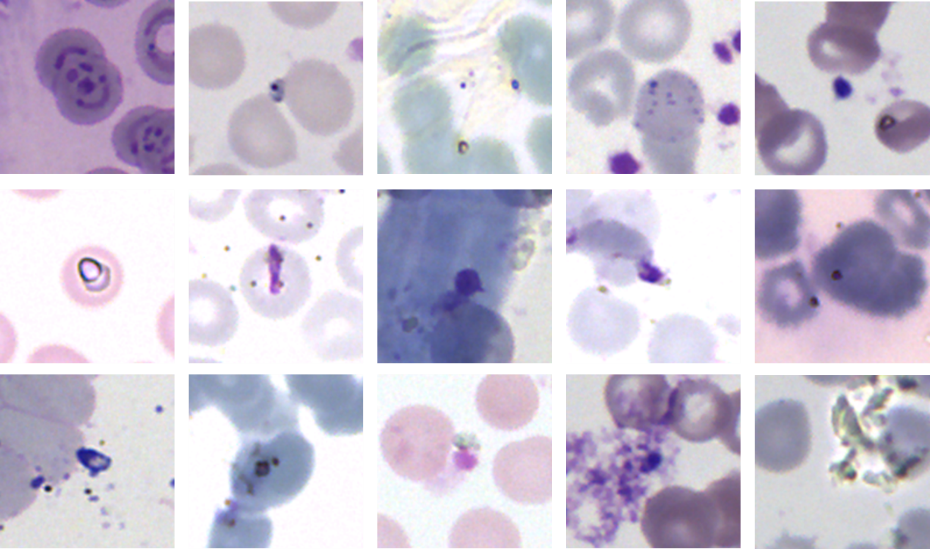}}
\caption{Distractors from field-prepared thin blood films. }
\label{distractors}
\end{center}
\end{figure*}

The malaria parasite has various developmental stages during its human blood life-cycle \cite{whoMicroscopy}. 
In the early stages after entering the blood stream, the parasites are in the ring stage (immature trophozoite), examples of which are shown in Fig. \ref{allStages} and \ref{rings}. 
Later stages of development include trophozoite, schizont, and gametocyte (sexual reproductive form). 
We collectively refer to these as ``late stage''. 
Example late stage parasites from the four species are shown in Fig. \ref{allStages}. 
In the ring stage, there are few differences between the various species of malaria, and they cannot be reliably distinguished by eye. 
They start to differentiate in subsequent stages of development (trophozoite, schizont, and gametocyte forms) which have marked distinctions between the different species.
An automated system must also differentiate between actual parasites  and distractor objects that resemble parasites. Some examples of distractors are shown in Fig. \ref{distractors}.  
  
\subsection{Thin blood film}
Malaria can be diagnosed with two types of blood films: 
thick film and thin film. 
Thick film allows for a larger volume of blood to be examined and thus provides a lower the limit of detection in terms of parasites per microliter ($\mu$L) of blood.
A calculation using the Poisson distribution (see section \ref{poissonSection}) indicates that, at a limit of detection (LoD) = 50 p/$\mu$L one must examine about 0.1 $\mu$L of blood to be fairly certain of at least one parasite being present.
This corresponds to 800 White Blood Cells (WBCs), which is easy to do on thick film; or 5e5 Red Blood Cells (RBCs) on thin film, which is unworkable since thin films are (ideally) a monolayer and also have large unusable regions.
 
Thus thick film is used for diagnosis. 
However,  distinguishing the three species \textit{P. vivax, P. ovale} and \textit{P. malariae} is very difficult  on thick films.
The thin film preparation preserves the morphology of parasites and RBCs, thus permitting species identification (ID). 

It is too time-consuming to scan a large volume of blood on thin film, making it less useful for quantitating low parasitemia samples. 
Thin films are used for quantitation when the parasite load is high because the number of parasites per field-of-view (FoV) is more manageable.

A good thin film slide is difficult to prepare. 
One places a drop of blood on a microscope slide, then uses another slide to spread the drop across the slide by capillary action. 
The RBCs on the edge of the blood film form a monolayer where distinct RBCs can be seen. 
The slide is then dried, fixed and stained with a Romanowsky-type stain, such as Giemsa. 
Due to protocols in different labs, stain pH level, and technician skill the background color of the thin film varied as observed in Figs. \ref{allStages}  - \ref{distractors}.
 
Under field conditions, an automated   system must handle color variation of background and RBCs, blurriness, out-of-focus images, overly-clumped RBCs, and distractors. 

\subsection{Distractor filter manual features} 
Manual features included area, intensity, extent, Euler number, eccentricity, mean gradient, prominence, bumpiness, roundness, ridgeness, donut shapeness, external contrast, internal contrast, mean of red, green and blue channels, coefficient of variation of gray scale, gradient, red, green and blue channels.

\subsection{Species ID module manual features}
The manual features included standard region properties such as elongation, perimeter, major and minor axis length, histogram of the grayscale image, pixel statistics of the gray-scale image; and properties of the gray-level co-occurrence matrix \cite{haralick} such as contrast, correlation, energy and homogeneity.

\subsection{Method for relaxing the annotation task}
Annotating ground truth on large datasets is expensive and time-consuming.
We were able to relax our parasite vetting task as follows:
when assessing algorithm performance (see ``Important metrics'' section) on the validation set, we used false positive (FP) rates from negative samples only.
Missed (i.e. unannotated) parasites were treated as distractors by the algorithm.
But  if the algorithm detected and classified these objects as parasites they counted as FPs and were thus disregarded, since they were on positive  samples.
The relevant metric on positive validation samples was sensitivity, which by definition considered only annotated parasites.
So while distractors mislabeled as parasites were harmful, missed parasites (on validation samples only) did not affect algorithm evaluation.
This method allowed imperfect annotations of validation samples, and thus allowed us to focus annotation resources on the   more important group of training set samples.

\subsection{Poisson variability and irreducible quantitation error} \label{poissonSection}
Rare events are governed by the Poisson distribution:\\

$P(k \text{ events in } N \text{ draws}) = e^{-pN} \frac{(pN)^k}{k!} $ \\\\ 
where $p$ = the probability of an event in one draw, and $N = $ number of draws.\\

This can be thought of as the limit of the binomial distribution $B(p,N)$:\\
 
$~~B(k,p,N) = P(k \text{ events } | ~p, N) = {N \choose k} p^k(1-p)^{N-k}$ \\ \\
as $p \rightarrow 0, N \rightarrow \infty$. This pushes the binomial probability mass function up against 0, i.e. $P(k \text{ events})$ becomes asymmetrical,  with $P(k \text{ events})$  highest for small $k$.

Let the parasitemia = $P$ parasites/$\mu$L.
Let $p$ be the probability that a particular RBC contains a parasite. Then $p = \frac{P}{5e6}$, assuming 5e6 RBCs/$\mu$L 
and at most one ring in any RBC (i.e. ignore the case of multiple rings in one RBC).
Consider each RBC as a coin toss with likelihood $p$ of coming up as ``parasite''. 
Then the total number of actual parasites in the RBCs examined is a binomial distribution $B(p, N)$. 
This implies that even a perfect annotator will count different numbers of parasites in different groups of RBCs from the same sample.
The variation depends on $P$ and $N$, and decreases as $P$ and/or $N$ increase. 
A similar situation holds for thick film counts, where $p$  the probability that a particular volume of blood (corresponding to one WBC) contains a parasite. Then $p = \frac{P}{8000}$, assuming 8000 WBCs/$\mu$L, a ``coin toss'' is examining 1/8000 $\mu$L of blood, and $N$ = number of WBCs counted (as a proxy for this blood volume).

We can quantify the amount of this ``irreducible Poisson error'' in quantitation using the relative standard error \\ \\
$~~~~relErr(P,N) = \frac{\text{std dev}(B(p,N))}{ \text{mean}(B(p,N))} = \frac{\sigma(B(p,N))} {\mu(B(p,N))}$.\\

In thick films, microscopists typically count 500  WBCs, while an automated scanner plus algorithm counts $\sim$1000 or 2000 WBCS. 
Values of $Q(P,N)$ for $N$ = 500, 1000, and 2000 are given in Fig. \ref{thickPoisson}. 
The advantage of machines, due to their ability to scan larger areas, is clear.

Microscopists switch to thin films for quantitation at around $P$ = 8k or 16k due to the difficulty of keeping track of counts on thick films. 
On thin film, microscopists typically count 1000 RBCs, while an automated scanner plus algorithm can count 10k to 20k RBCs. 
Values of $Q(P,N)$ for $N$ = 1k, 10k, 20k are given in Fig. \ref{thinPoisson}, again showing the clear advantage of machines.
Machines have an additional advantage due to their ability to accurately count parasites on thick films at parasitemias up to 80k/uL. This drastically reduces Poisson error vs microscopists in the 16k $< P <$ 80k range, because of the much larger volume of blood examined on thick vs thin film. 
This is seen in Fig. \ref{combinedPoisson}, which combines Figs \ref{thickPoisson} and \ref{thinPoisson} and plots relative standard error for thick and thin film quantitation, at all parasitemias and for a variety of WBC and RBC counts.

\section{CNN architectures} 
The two CNN architectures (ring and late stage branches) are shown in Fig. \ref{cnnSchematics}.
The  ring branch network has multi-scale convolutional kernels inspired by the Inception \cite{szegedy} architecture.  
The late stage branch network is a fully convolutional architecture with strided convolutions to achieve size reduction. 
Each convolutional layer also includes ReLU, and convolutional layers 2-7 are followed by dropout (0.3\%-0.5\%).
%The late stage branch CNN architecture is shown in Fig \ref{lateCnn}.
%ADD CONTENT AS NEEDED. 

%\begin{figure*}[h!]
%\begin{center}
%\centerline{
%\includegraphics[width=0.8\linewidth]{lateCnnSchematic.png}}
%\caption{Late stage branch CNN architecture}
%\label{lateCnn}
%\end{center}
%\end{figure*} 

\subsection{Output Report}
\begin{figure*}[h!]
\begin{center}
\centerline{
\includegraphics[width=\linewidth]{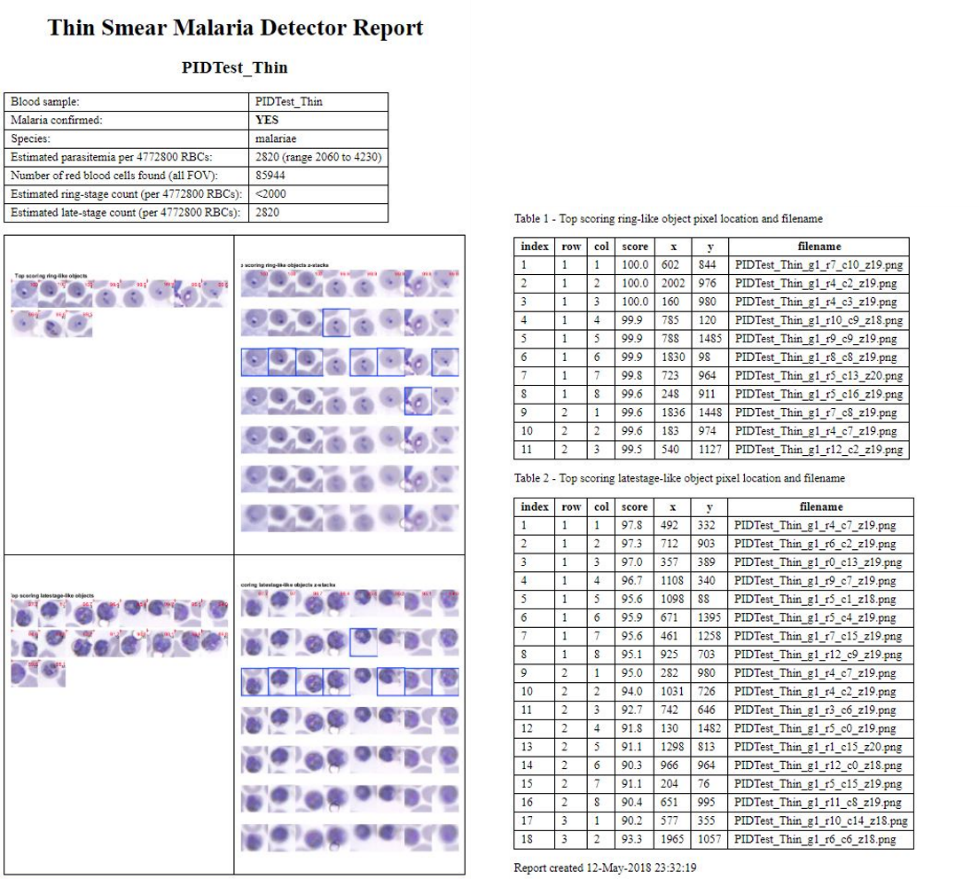}}
\caption{Report 1. }
\label{report1}
\end{center}
\end{figure*}

\begin{figure*}[t!]
\begin{center}
\centerline{
\includegraphics[width=\linewidth]{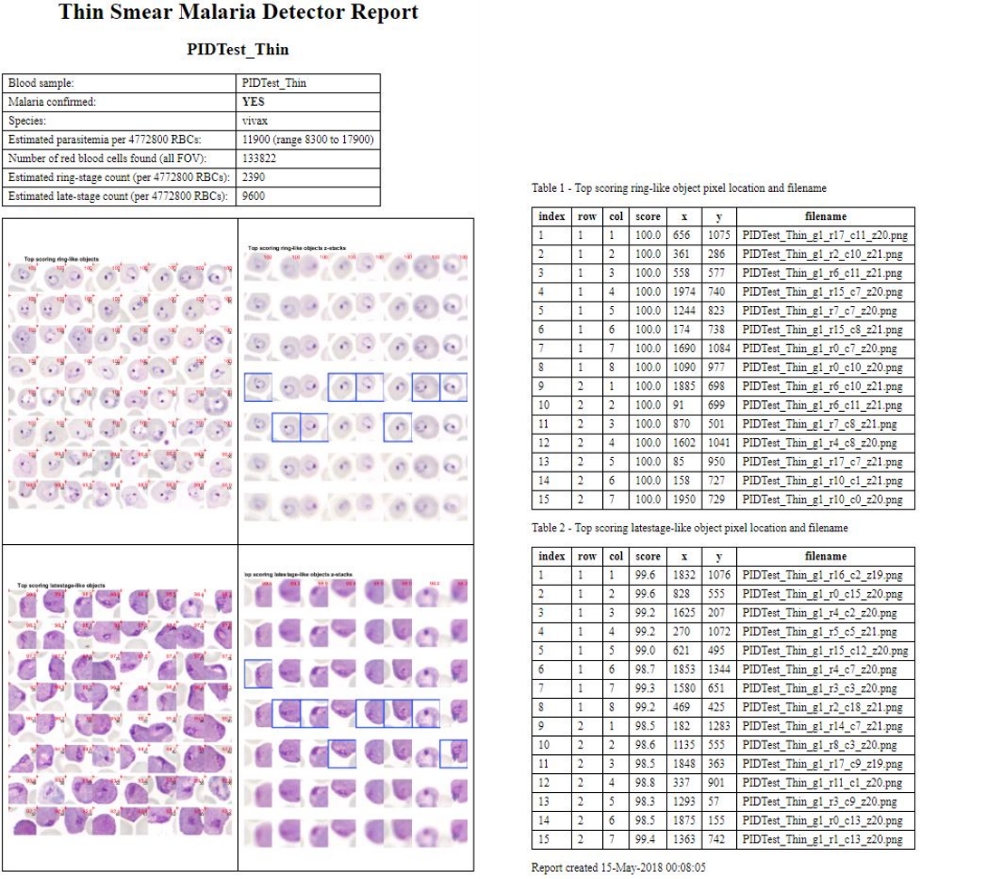}}
\caption{ Report 2. }
\label{report2}
\end{center}
\end{figure*}

 Two sample reports generated by our thin film malaria assessment system are shown in Figs. \ref{report1} and \ref{report2}. 
The report lists the detected malaria species, the ring and late-stage quantitations, and number of RBCs examined. 
It also shows a mosaic of the highest-scoring thumbnails from both the ring branch and the late stage branch.
These thumbnails can serve as a decision aid to a microscopist in low resource setting: it is a method of collecting objects-of-interest from a large region of blood film for visual examination.
It can also serve to reassure a trained technician that the algorithm (nominally a ``black box'') is delivering reasonable results.\\  

\begin{figure*} [h!]
\begin{center}
\centerline{
\includegraphics[width=\linewidth]{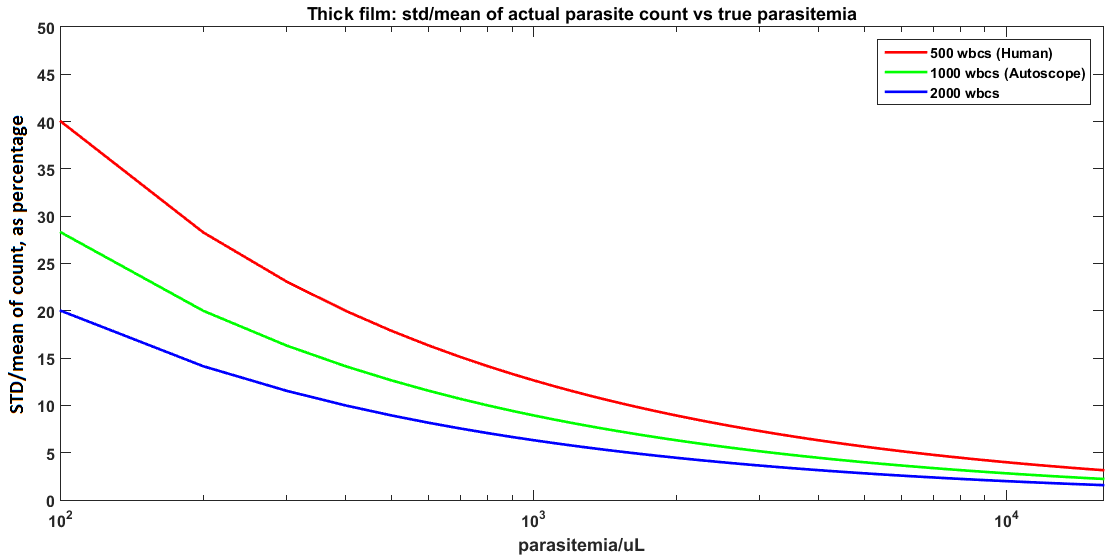}}
\caption{Irreducible Poisson error on thick films for P from 10/$\mu$L to 16k/$\mu$L, at three values of $N$ = \# WBCs counted:
 500 (red curve), 1000 (green curve), and 2000 (blue curve).
Microscopists typically examine blood volume containing 500 WBCs.
Automated devices and algorithms can conveniently scan 1000 - 2000 WBCs' worth of blood, reducing Poisson variation.}
\label{thickPoisson}
\end{center}
\end{figure*}

\begin{figure*} [h!]
\begin{center}
\centerline{
\includegraphics[width=\linewidth]{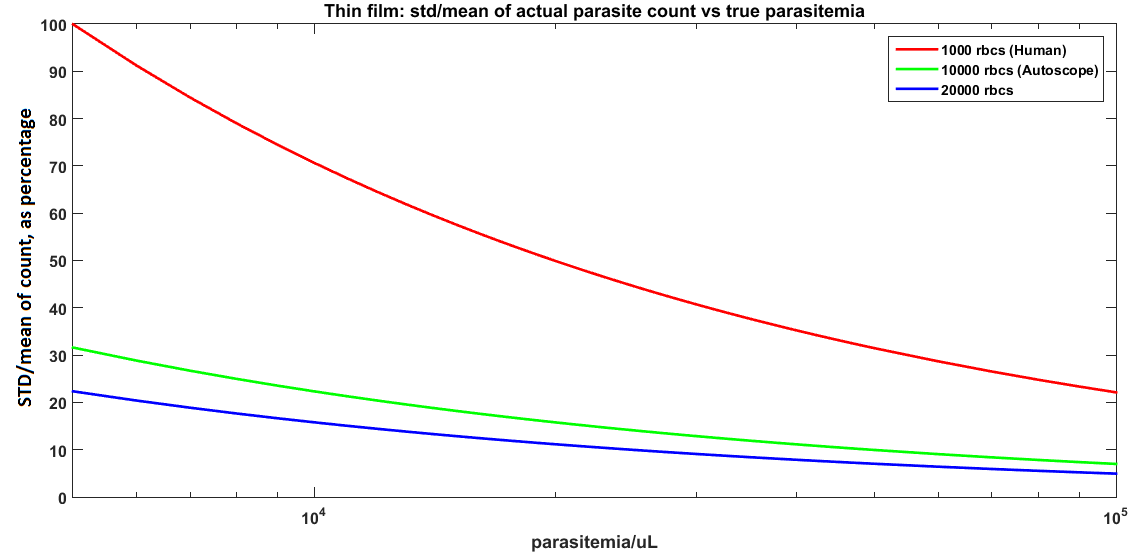}}
\caption{ Irreducible Poisson error on thin films, for 5k $< P <$ 100k/$\mu$L, , at three values of $N$ = \# RBCs counted: 
1000 (red curve), 10,000 (green curve), and 20,000 (blue curve).
Microscopists typically examine blood volume containing 1000 RBCs.
Automated devices and algorithms can conveniently scan 10,000 - 20,000 WBCs' worth of blood, substantially reducing Poisson variation.}
\label{thinPoisson}
\end{center}
\end{figure*}

\begin{figure*} [h! ]
\begin{center}
\centerline{
\includegraphics[width=\linewidth]{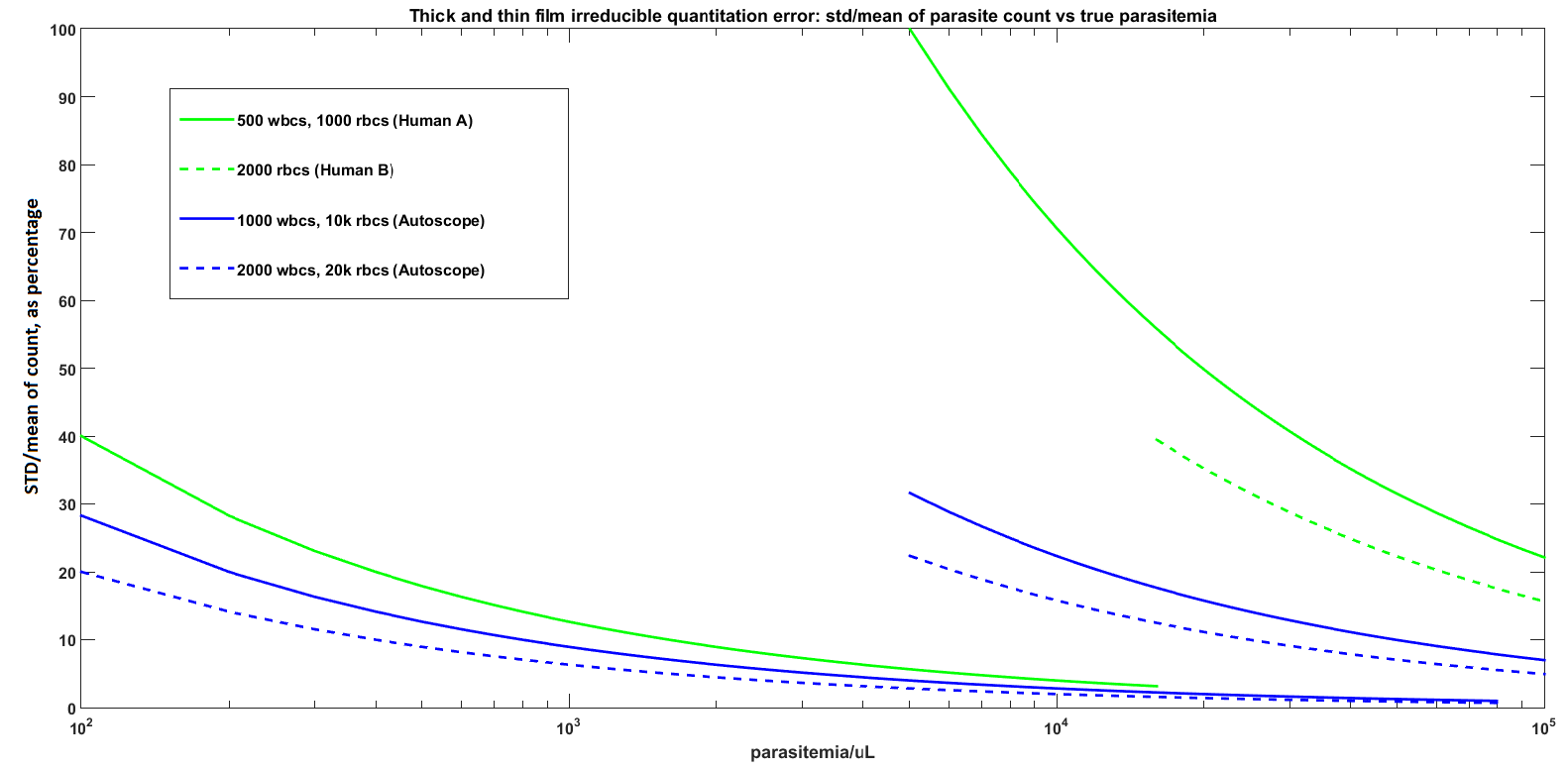}}
\caption{Irreducible Poisson error for all parasitemias, for various numbers of WBCs or RBCs examined. The two top pairs of lines, that start at P = 5k/$\mu$L (or P = 16k/$\mu$L), correspond to RBC counts on thin film. The three lower lines correspond to WBC counts on thick film.}
\label{combinedPoisson}
\end{center}
\end{figure*}

\begin{figure*}[h!]
\begin{center}
\centerline{
\includegraphics[width=0.6\linewidth]{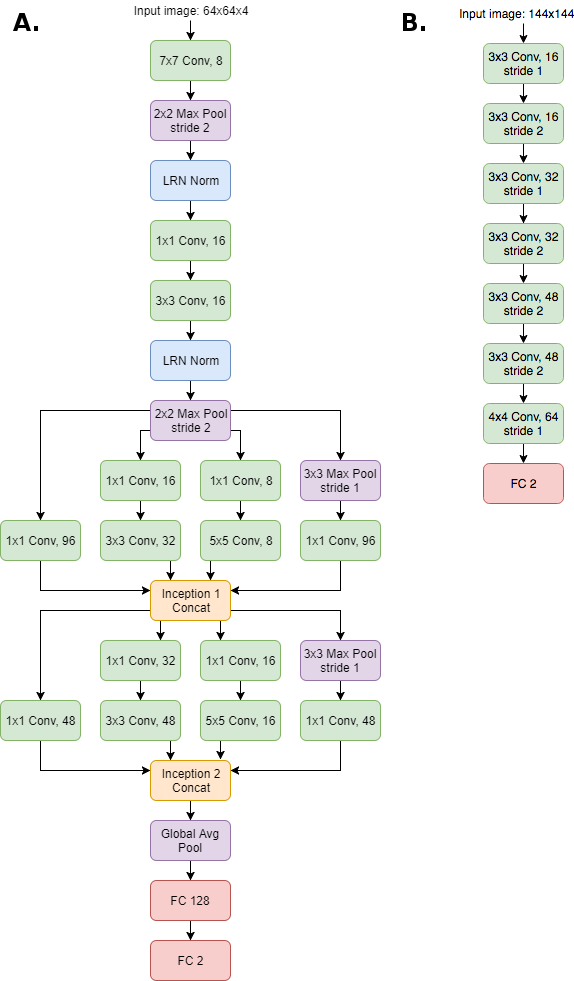}}
\caption{CNN architectures. \textbf{A:} Ring branch. \textbf{B:} Late stage branch.}
\label{cnnSchematics}
\end{center}
\end{figure*}

Figs \ref{thickPoisson} to \ref{report2} follow the References section.\\

% The rest of the S.I. figures:

\end{document}